\documentclass[11pt]{article}

\usepackage[preprint]{acl}

\usepackage{times}
\usepackage{latexsym}

\usepackage[T1]{fontenc}

\usepackage[utf8]{inputenc}

\usepackage{microtype}

\usepackage{inconsolata}

\usepackage{graphicx}
\usepackage{amsmath}
\usepackage{tcolorbox}
\usepackage{multicol}
\usepackage{cuted} 
\usepackage{longtable}
\usepackage{booktabs}  

\usepackage{fancyhdr}
\title{Ro-SLM: Onboard Small Language Models for Robot Task Planning and Operation Code Generation}

\author{
 \textbf{Wenhao Wang\textsuperscript{1}},
 \textbf{Yanyan Li\textsuperscript{2}},
 \textbf{Long Jiao\textsuperscript{1}},
 \textbf{Jiawei Yuan\textsuperscript{1}}
\\
\\
 \textsuperscript{1}Department of Computer \& Information Science, University of Massachusetts Dartmouth
 \\
 \textsuperscript{2}Department of Computer Science \& Engineering, California State University San Marcos 
\\
{\tt\small \{wwang5, ljiao, jyuan\}@umassd.edu, yali@csusm.edu}
\\
 }

\begin{document}
\maketitle
\begin{abstract}
Recent advances in large language models (LLMs) provide robots with contextual reasoning abilities to comprehend human instructions. Yet, current LLM-enabled robots typically depend on cloud-based models or high-performance computing infrastructure, which limit their deployment on robots under unreliable internet environments or with constrained computational resources, such as UAVs and small ground vehicles. Thus, deploying fine-tuned small language models (SLMs) that support onboard deployment offers a promising alternative. This paper introduces Ro-SLM, a framework that enables reliable SLM-driven robot operation by distilling LLMs' knowledge and reasoning. Ro-SLM starts from dataset synthesis by leveraging LLMs to generate diverse task instructions, produce corresponding ground truth code with minimal human assistance, and augment instructions into real-world application scenarios. Ro-SLM is then fine-tuned with the dataset, in which LLM serves as a reward function to guide the training. Extensive experiments on UAV operation tasks demonstrate that Ro-SLM improves the performance of SLM from being incapable of supporting robotic task planning and code generation to achieving performance that approaches LLM\footnote{Project is available at \url{https://github.com/ai-uavsec/Ro-SLM}}.

\end{abstract}

\section{Introduction} \label{sec:intro}

Recent advances in LLMs~\cite{GPT4, Gemini} have demonstrated remarkable proficiency in robotic areas such as control~\cite{Real}, planning~\cite{planning}, and navigation~\cite{L3MVN}. Trained on internet-scale data and comprising billions of parameters, LLMs encode rich knowledge and exhibit strong semantic understanding and context-generation capabilities, enabling robots to reliably comprehend human instructions for planning and generate corresponding robot operation code~\cite{ChatGPTRobotics, CodeasPolicy, gemini-robotics}. As a result, increasing efforts have focused on designing structured prompting strategies~\cite{PromptEngineering} to enhance foundational LLMs (e.g., OpenAI GPT and Gemini) in supporting robotic tasks reliably~\cite{PromptBook, CodeasPolicy, GSCE}.


\begin{figure}[t]
    \centering
    \includegraphics[width=1\linewidth]{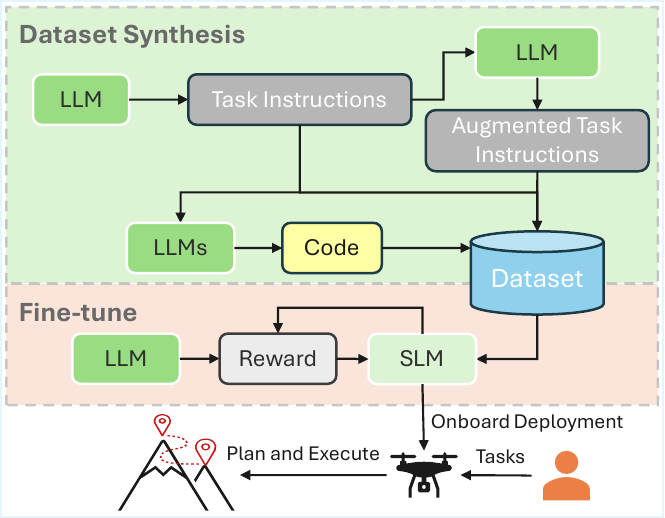}
    \caption{Ro-SLM framework for enabling SLM-driven robot operations.}
    \label{fig:overall}
\end{figure}

Despite LLMs' strong reasoning capabilities, the scale of LLMs brings significant overhead and limits their applications on resource-constrained robotic platforms. On the one hand, cloud-based LLMs such as OpenAI GPT4~\cite{GPT4} require a stable internet connection to access remote inference services, which is infeasible for robots operating in environments with unreliable connectivity, such as mountainous regions or disaster response. On the other hand, locally deploying LLMs (e.g., Llama-3.1-405B~\cite{llama3}) demands significant computational and energy resources, making them impractical for robots with limited payload and battery capacity, such as small ground vehicles or UAVs. Leveraging SLMs for resource-constrained robotic platforms provides another promising alternative ~\cite{SLM-survey}. While the deployment of SLMs has been practically demonstrated, they exhibit weaker reasoning capabilities due to their reduced parameter scale~\cite{LLM-scaling}, which hence limits their ability to extract task-relevant information from prompts and generalize to novel robotic tasks like LLMs.

To overcome these limitations, we propose Ro-SLM (Robot operation with Small Language Model), a framework that enables reliable SLM-driven robot operations by distilling the knowledge and reasoning from LLMs. As illustrated in Figure~\ref{fig:overall}, Ro-SLM consists of two stages:


\textbf{Dataset Synthesis} A critical component of effective fine-tuning of Ro-SLM is constructing a high-quality dataset that consists of varying task complexity scenarios and corresponding execution plans and operation code. Given the challenge that the generation of such robot task and execution data is time-consuming and requires domain experts, this paper configured LLMs with expert knowledge to facilitate the construction and validation of a high-quality synthetic fine-tune dataset. Specifically, an instruction set is first generated by configuring an LLM agent to automatically generate task instructions with varying complexities and clearly specified task objectives, thereby reducing the risk of misinterpretations during code generation. Then, a corrective code generation approach~\cite{LLM-simulator} accurately generates corresponding ground truth robot operation code with human assistance introduced only for high-complexity tasks. To align SLM with real human-like task descriptions, we further leverage the world knowledge of LLM to augment the instruction set with real-world robotic application scenarios. In addition, to prevent the SLM from overfitting to syntactic patterns in code comments, we remove all comments from the ground truth code set.

\textbf{Fine-tune} After constructing the dataset, the SLM is first trained using supervised fine-tuning (SFT). To ensure the model is getting the best result, the LLM then serves as a reward function to guide the model optimization process.


We conducted extensive experiments on UAV operation tasks with varying levels of complexity to evaluate the effectiveness of Ro-SLM. Our evaluation shows that the Ro-SLM with Llama-3.1-8B achieves $97.7\%$ Success Rate and $98.9\%$ Completeness on simple tasks, i.e., $98.9\%$ of required actions in all evaluated tasks are completed correctly and $97.7\%$ of tasks are entirely completed without any error. For complex tasks, Ro-SLM achieves a $70\%$ Success Rate and $83.7\%$ Completeness. Compared with the performance of OpenAI o4-mini LLM, Ro-SLM achieves a comparable performance. Thus, Ro-SLM is promising to act as an intelligent agent for resource-constrained robot platforms for task planning and operation code generation.

\begin{figure*}[t]
    \centering{\includegraphics[width=0.975\textwidth]{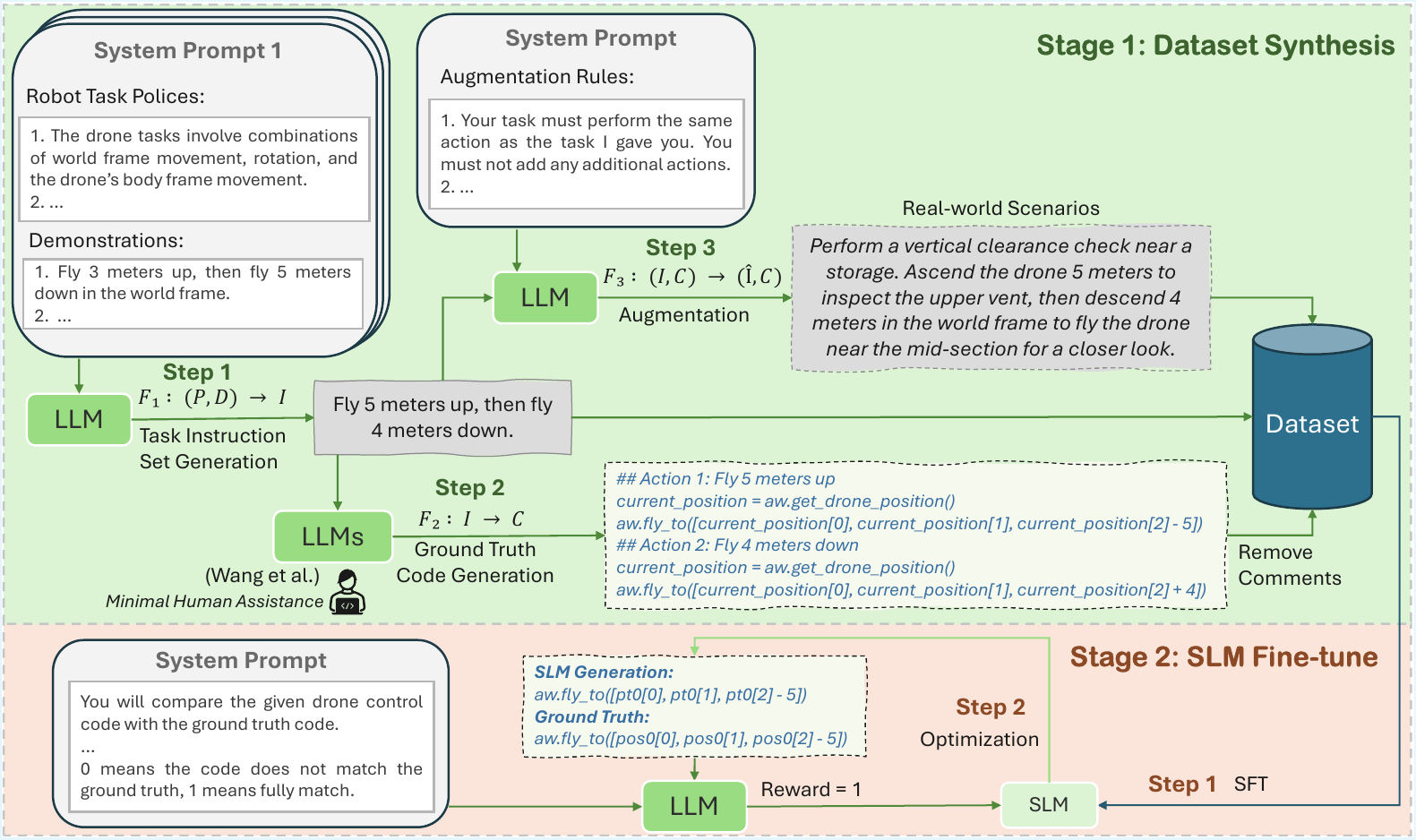}}
    \caption{Ro-SLM overview: dataset synthesis and SLM fine-tuning. The LLMs are configured with different system prompts to support instruction generation, augmentation, ground truth code mapping, and the reward function. The detailed system prompts are provided in Appendix~\ref{apdx:instruction sysprompt}, \ref{apdx:aug sysprompt}, and \ref{apdx:reward sysprompt}.}
    \label{fig:data gen}
\end{figure*}

\section{Related Work} \label{sec:related work}

\subsection{LLM-driven Robot Code Generation}

Compared with traditional pipelines in which human experts manually craft robot operation code from task instructions, recent research has increasingly leveraged LLMs' natural language understanding and reasoning capabilities to automatically infer tasks and generate corresponding robot-operation code or plans~\cite{GSCE, ChatGPTRobotics, TypeFly, CodeasPolicy}. To enhance reasoning performance and ensure the correctness of generated code for robotic tasks that require complex reasoning, prior works have employed prompt engineering~\cite{PromptEngineering} to shape LLM behavior through structured system prompts, such as constraints and examples~\cite{PromptBook, GSCE}. More recent approaches further improve reliability by adopting corrective code-generation frameworks, in which errors and mismatches in generated code are identified and iteratively refined~\cite{AutoTAMP, CoPAL, CLGSCE}. Notably, most existing methods rely on LLMs (e.g., OpenAI GPT) due to their remarkable reasoning capabilities; however, these methods depend on cloud-based computation and internet connectivity and therefore cannot be deployed onboard robots. In contrast, SLMs have significantly fewer parameters and are more suitable for onboard deployment, but they struggle to achieve comparable performance as the LLMs~\cite{roboinstruct}.

\subsection{Dataset Generation}

Fine-tuned SLMs can achieve performance comparable to that of LLMs on domain-specific tasks~\cite{zhan2025slm}. However, unlike domains such as computer vision, robot tasks are customized and robot capabilities-dependent, resulting in fine-tuning dataset construction being labor-intensive and difficult to scale. Recent approaches have leveraged LLMs' world knowledge acquired from the large-scale pretraining to synthesize a dataset for fine-tuning. For example, LLM-Trainer uses LLMs to create and annotate robot trajectories based on the demonstrations~\cite{LLM-Trainer}. PRISM distills an LLM-enabled planner to train an SLM that rivals the planning performance of LLMs~\cite{Distill}. Other work~\cite{Small-VLA, Small-nav, OFSLMs} incorporates human-in-the-loop strategies for synthesizing task instructions and robot operation plans. Moreover, LLMs are prone to hallucinations and SLMs are sensitive to data contamination~\cite{SLM-sensitivity}. Therefore, ensuring the quality of dataset generation is crucial for the reliability of fine-tuned SLMs.



\subsection{SLM Fine-tune}
Fine-tuning adapts pretrained language models to specific tasks by training on labeled input-output pairs so that the model learns to follow desired behaviors or instructions~\cite{llm-finetune}. Several techniques have been proposed to improve the efficiency of fine-tuning. To reduce the GPU memory usage, LoRA introduces low-rank trainable updates to pretrained model weights, which significantly reduces the number of trainable parameters and the size of optimizer states~\cite{lora}. Moreover, to align model outputs with human preferences, reinforcement learning from human feedback (RLHF) optimizes the model using rewards derived from human feedback~\cite{rlhf}. To further improve the efficiency of RLHF, group relative policy optimization (GRPO)~\cite{grpo} removes the need for training a reward model as used in PPO-based~\cite{PPO} approaches, which also reduces memory usage and accelerates training.


\section{Method} \label{sec:method}

\subsection{Overview}

Figure~\ref{fig:data gen} presents the workflow of our Ro-SLM framework. In the \underline{\emph{Stage 1}}, we synthesize the fine-tune dataset, which consists of three key steps: (1) \textbf{Task Instruction Set Generation}: an LLM-based agent generates a set of concise and executable task instructions for robotic operations of varying complexity; (2) \textbf{Ground Truth Code Generation}: LLM agents with enhanced reasoning produce corresponding ground truth code with minimal human assistance; (3) \textbf{Augmentation}: an LLM agent augments the dataset such that the task instructions are aligned to real-world applications. In  \underline{\emph{Stage 2}}, with the dataset prepared, the SLM is initially trained using SFT and further optimized via GRPO with rewards provided by the LLM reward function. 


\subsection{Task Instruction Set Generation} \label{sec:task gen}

The goal of task instruction generation is to create a set of executable task instructions for robot operations. To achieve this, we leverage reasoning and the context generation capabilities of LLM to infer feasible robot operations and generate a set of task instructions. Formally, we view the LLM agent as implementing a function $\mathcal{F}_{1} : (P, D) \rightarrow I$, where $P$ denotes the robot operational tasks policies and $D$ is task instruction demonstrations. Given $(P, D)$, the LLM generates a set of task instructions \(I\). 

\textbf{Task Generation:} To prevent LLM hallucination that could produce non-executable tasks, we employ in-context learning to guide the LLM toward reliably producing valid task instructions. Specifically, we construct a system prompt to enhance a foundational LLM. First, we define the LLM's role of generating robot operational task instructions. Next, we provide the policies $P$ for robot operations and task instruction formats. These policies are essential for the LLM to understand the robot's capabilities and operational constraints. Moreover, the policies enforce the LLM to produce clear and concise instructions, thereby minimizing ambiguity that could lead to misinterpretation in the subsequent code generation. Finally, we construct $k$ demonstrations of task instructions, which cover the fundamental robot operation patterns and standardized instruction formats. The LLM learns to imitate and recombine these patterns while generating new task instructions. 

\textbf{Task Distribution:} After configuring the foundational LLM with the above structured system prompt, we prompt the LLM agent to generate $n$ task instructions. To increase diversity, we further design multiple system prompt configurations for varying task complexity distributions. As a result, the instruction set $I$ spans multiple levels of complexity, ranging from few-stage planning and basic spatial reasoning (one or two actions) to multi-stage complex reasoning (five or more actions require complex reasoning of the robot's dynamics). The diversity ensures that the SLM is fine-tuned on a broad task distribution, which enables the SLM to develop robust and generalizable reasoning capabilities across varying task complexity scenarios. 

\subsection{Ground Truth Code Generation} \label{sec:code gen}

To reduce human effort, we leverage the reasoning and planning capabilities of LLMs to ground each instruction in $I$ into code. Specifically, the LLM agents implement a grounding function $\mathcal{F}_{2}: I \rightarrow C$, which maps each task instruction in $I_i \in I$ to its corresponding ground truth code $C_{i}$. 

\textbf{LLM Code Generation:} To ensure reliable instruction grounding, we adopt the corrective code generation framework proposed in~\cite{LLM-simulator}, which identifies and corrects errors in the generated code before producing the final version $C_i$ for each task instruction $I_i$.

\textbf{Code Correctness Control:} While LLMs can reliably generate correct code for simple tasks, their performance degrades on tasks requiring complex spatial reasoning or long-horizon planning~\cite{CLGSCE, LLM-simulator}. Therefore, fully LLM-based grounding may introduce errors in $C$, leading to the SLM being fine-tuned on incorrect supervision. To ensure the correctness of $C$, we adopt a hybrid code generation strategy, in which human oversight is used for code generation for tasks that involve five or more steps of action. In these tasks, a human expert reviews the generated code and corrects any errors with respect to the task instructions. 


\subsection{Augmentation}

As discussed in Section~\ref{sec:task gen}, the instruction set $I$ must be clear and concise to prevent misinterpretation during subsequent code generation, which makes the task instructions expressed in high-level operational plans (e.g., \textit{``Fly 5 meters up, then fly 4 meters down.''}). However, such high-level plans usually differ from the way humans typically describe tasks in real-world settings, such as the real-world scenario example presented in Figure \ref{fig:data gen}. To bridge this gap, we augment the dataset by mapping the task instructions into robot application scenarios. Specifically, we leverage the world knowledge in LLMs to rephrase each instruction to a description of a corresponding real-world robot deployment. The augmentation function is defined as $\mathcal{F}_{3} : (I, C) \rightarrow (\hat{I}, C)$, where $(\hat{I}, C)$ represent the augmented instruction and corresponding code pairs.


\textbf{Augmentation Quality Control:} The primary challenge in dataset augmentation is to ensure that each augmented instruction $\hat{I_i}$ describes the same robot operations as the original instruction $I_i$, such that the code $C_i$ matches with the augmented tasks during SLM fine-tuning. To achieve this, we leverage in-context learning of LLMs again. We construct a system prompt that imposes rules on the task-mapping process to prevent introducing additional robot actions, and the $\hat{I}$ should be feasible real-world robot operations. Moreover, to avoid overly verbose and unnatural generations~\cite{LLM-style}, LLM is prompted to use common vocabulary, one paragraph description, and human-like tones in its augmentation.

\subsection{Dataset Construction}

We construct the final fine-tune dataset by applying $\mathcal{F}_{1}$, $\mathcal{F}_{2}$, $\mathcal{F}_{3}$ in sequence. The final dataset $S$ consists of paired instructions and code, denoted as $S = (I \cup \hat{I}, C)$. Furthermore, since the SLMs are sensitive to the syntactic patterns of fine-tuning data and their performance degrades under syntactic contamination~\cite{SLM-sensitivity}, we remove all comments from the code in the dataset. This data preprocessing strategy reduces the risk that the SLM overfits to linguistic patterns rather than the code, which ensures the fine-tuned SLM focuses on producing robot operational code when given a task instruction.

\subsection{SLM Fine-tune}

Given the dataset $S$, the fine-tune process trains an SLM $M$ to implement $\mathcal{F}: (M, I') \rightarrow C'$, such that $M$ generates robot operation code $C'$ when provided with a task instruction $I'$. The fine-tuning procedure consists of two steps:

\textbf{Supervised Fine-Tuning}: The $M$ is fine-tuned on the dataset $S$ to align the SLM with the desired code generation style and enables it to learn robot actions (i.e., APIs) and their mappings from task instructions to executable code. Additionally, we employ LoRA~\cite{lora} to improve fine-tuning efficiency and reduce memory usage.


\textbf{Optimization with LLM Feedback}: In the second stage, we further optimize $M$ to enhance step-level reasoning by GRPO~\cite{grpo} with LLM as reward function. Unlike text-generation settings, where being semantically similar to the ground truth is often sufficient for a high reward, in robotic code generation, even semantically similar code may implement different robot actions. Conversely, semantically dissimilar expressions may correspond to the same underlying operation (e.g., variable ``position'' and ``base'' both referring to the robot's location). To address these issues, we configure an LLM to interpret the robot behavior implied by the code generated by the SLM and compare it with the ground truth. The LLM then outputs feedback indicating whether the generated code executes the same robot actions as the ground truth. This feedback is then constructed as a reward for GRPO to optimize the SLM's weights.


\section{Experiments} \label{sec:experiment}

\subsection{Experimental Setup}
\subsubsection{Implementation}
We implement the dataset synthesis in our proposed Ro-SLM framework using OpenAI GPT models. Specifically, task instruction generation and augmentation are performed with GPT-5.1~\cite{gpt5.1}, while ground truth code generation is conducted using o4-mini~\cite{o4mini}. For the training set, we synthesize $203$ task instructions, of which $150$ are automatically grounded into code, and $53$ require human assistance for grounding. Dataset augmentation results in a final dataset that consists of $598$ instruction-code pairs. The dataset is then split into $492$ pairs for training $T$ and $106$ pairs for evaluation $E$.

The fine-tuning of Ro-SLM is implemented with Llama-3.1-8B~\cite{llama3} from Hugging Face~\cite{huggingface}, which can be deployed for inference on small robots with their onboard processing units (e.g., Jetson AGX Orin). In the optimization, the reward function is implemented by GPT-5.1~\cite{gpt5.1}. Both training and optimization of Ro-SLM are conducted on a server with two Nvidia A100 GPUs.

 We use the pre-built ``block''~\cite{Blocks} environment of the AirSim simulator~\cite{AirSim}. All results are averaged over three runs to mitigate the randomness of LLM generation~\cite{LLMrandomness}.

\subsubsection{Baselines}
Three methods are selected as the baselines for comparison:

\textbf{GSCE}~\cite{GSCE}: This method configured a foundational o4-mini with a structured system prompt, enabling the LLM to generate UAV operation code according to human instructions. 

\textbf{Corrective GSCE}~\cite{LLM-simulator}: The state-of-the-art (SOTA) method for robot operation code generation. Build upon GSCE; the code is iteratively refined towards achieving the task objectives, which is also the implementation of ground truth code generation in section~\ref{sec:code gen}. Notably, all the LLMs in this method use o4-mini. 

\textbf{Llama-3.1-8B}: This baseline uses the original SLM without fine-tuning. The original Llama-3.1-8B~\cite{llama3} downloaded from Huggingface is configured with the same structured system prompt as GSCE~\cite{GSCE} to act as an agent for UAV operation code generation, which has been proven effective on LLMs.

\subsubsection{Test Datasets}
The Basic and Advanced datasets from~\cite{CLGSCE} are selected as benchmark datasets for a comprehensive evaluation of the model's reasoning capabilities across varying levels of task complexity. The Basic dataset consists of 44 intuitive UAV operation task instructions and each involves one or two actions. They are designed to assess a few-stage planning and basic logical reasoning. The Advanced dataset contains $20$ long-horizon UAV operation tasks with four levels of complexity (``\textit{Low}'', ``\textit{Medium}'', ``\textit{High}'', and ``\textit{Very High}''). Each advanced task requires complex reasoning about UAV dynamics in the geometric world in order to make correct decisions.  Notably, all tasks in both datasets are manually validated to prevent potential simulation-induced errors that could affect the experiment result. 

The above tasks evaluate the model’s ability to reason about task logic and follow instructions. To assess robustness, we further extend both datasets by mapping the original instructions to real-world application scenarios, which assesses whether the models are able to reliably respond to task instructions that are more natural and practical. The extended datasets are provided in Appendix~\ref{apdx:modified dataset}.

\begin{table*}[t]
  \centering
  \begin{tabular}{lcccc}
    \toprule
    & \multicolumn{2}{c}{\textbf{Basic}} & \multicolumn{2}{c}{\textbf{Advanced}}\\
    \midrule
    Method & SR & Completeness & SR & Completeness \\
    \midrule
    Llama-3.1-8B~\cite{llama3} & $ 9.1\%$ & $ 9.1\%$ & $ 5\%$ & $ 9\%$ \\
    GSCE~\cite{GSCE} & $ 100\%$ & $ 100\%$ & $ 75.0\%$ & $ 91.5\%$  \\
    Corrective GSCE~\cite{LLM-simulator} & $ 100\%$ & $ 100\%$ & $ 90.0\%$ & $ 97.8\%$  \\
    
    Ro-SLM with Comments & $ 95.5\%$ & $ 96.6\%$ & $ 50.0\%$ & $ 76.2\%$  \\
    Ro-SLM without Augmentation & $ 95.5\%$ & $ 96.6\%$ & $ 55.0\%$ & $ 74.8\%$  \\

    Ro-SLM Simple Task Only & $ 90.9\%$ & $ 93.2\%$ & $ 0\%$ & $ 1.9\%$ \\
    Ro-SLM Complex Task Only & $ 0\%$ & $ 2.3\%$ & $ 75\%$ & $ 83.8\%$ \\

    Ro-SLM without GRPO Optimization & $ 97.7\%$ & $ 98.9\%$ & $ 70.0\%$ & $ 83.1\%$ \\

    Ro-SLM & $ 97.7\%$ & $ 98.9\%$ & $ 70.0\%$ & $ 83.7\%$ \\
    \bottomrule
  \end{tabular}
  \caption{Results on the Basic and Advanced datasets.}
  \label{tab:overall}
\end{table*}

\begin{table*}[t]
  \centering
  \begin{tabular}{lcccc}
    \toprule
    & \multicolumn{2}{c}{\textbf{Basic}} & \multicolumn{2}{c}{\textbf{Advanced}}\\
    \midrule
    Method & SR & Completeness & SR & Completeness \\
    \midrule
    Llama-3.1-8B~\cite{llama3} & $ 11.4\%$ & $ 26.1\%$ & $ 0\%$ & $ 8.4\%$  \\
    GSCE~\cite{GSCE} & $ 100\%$ & $ 100\%$ & $ 75.0\%$ & $ 94.8\%$  \\    
    Corrective GSCE~\cite{LLM-simulator} & $ 100\%$ & $ 100\%$ & $ 81.7\%$ & $ 96.3\%$  \\
    
    Ro-SLM with Comments & $ 90.9\%$ & $ 93.2\%$ & $ 55.0\%$ & $ 75.2\%$  \\
    Ro-SLM without Augmentation & $ 88.6\%$ & $ 92.0\%$ & $ 70.0\%$ & $ 81.3\%$  \\

    Ro-SLM Simple Task Only & $ 90.9\%$ & $ 94.3\%$ &  $ 0\%$ & $ 3.6\%$  \\
    Ro-SLM Complex Task Only & $ 0\%$ & $ 2.3\%$ & $ 70\%$ & $ 83.0\%$ \\

    Ro-SLM without GRPO Optimization & $ 93.2\%$ & $ 95.5\%$ & $75\%$ & $ 86.4\%$ \\
    
    Ro-SLM & $ 93.2\%$ & $ 95.5\%$ & $75\%$ & $ 87.0\%$ \\
    \bottomrule
  \end{tabular}
  \caption{Results on the real-world mapping of Basic and Advanced datasets.}
  \label{tab:overall modified}
\end{table*}

\subsubsection{Evaluation Metrics}
We evaluate performance using \textbf{Completeness} and \textbf{Success Rate (SR)}~\cite{CLGSCE}. The ground truth is represented by a sequence of UAV state transitions. Each state transition is a vector of four elements: \([x, y, z, \theta]\), where \(x\), \(y\), and \(z\) denote the robot's position transitions in the North, East, and Down axes, and \(\theta\) represents the yaw rotation.

\textbf{Completeness} measures the proportion of actions that are executed correctly in a given task. Given a task \(i\), its Completeness is defined as Eq.~\ref{eq:completeness_i}, i.e., the ratio between the number of correctly executed actions (\(a^{\mathrm{correct}}_i\)) and the total number of ground truth actions (\(l^{\mathrm{gt}}_i\)) in the task. This metric provides insight into reasoning performance throughout the intermediate code execution process.


\begin{equation}
    \mathrm{Completeness}_i = \frac{|a^{\mathrm{correct}}_i|}{|l^{\mathrm{gt}}_i|}
    \label{eq:completeness_i}
\end{equation}

\textbf{SR} captures task-level performance by measuring whether the robot successfully reaches the final goal state while correctly following intermediate actions. Given a task $i$, it is considered successful ($SR_i$=1) if and only if the entire action sequence is executed without error, i.e., $Completeness_i = 1$; otherwise, $SR_i$=0.





\subsection{Results}

\subsubsection{Overall Result}


The overall results presented in Table~\ref{tab:overall} indicate that the SLM fine-tuned by Ro-SLM achieves comparable performance to the LLM. Specifically, Ro-SLM obtains $98.9\%$ Completeness and $97.7\%$ SR for the Basic tasks; $83.7\%$ Completeness and $70.0\%$ SR for the Advanced tasks, which are close to the performance achieved by GSCE. In contrast, applying the same prompting strategies as GSCE to the original Llama-3.1-8B is incapable of producing accurate robot operation code; in our experiment, the model either repeats the task instructions in its generation or produces incorrectly reasoned code. 

Moving on to the real-world instruction mapping results in Table~\ref{tab:overall modified}. On the Basic tasks, the Ro-SLM achieves $95.5\%$ Completeness and $93.2\%$ SR. For the Advanced tasks, Ro-SLM shows improvement; it obtains $87\%$ Completeness and $75\%$ SR, which reduces the performance gap with the LLM on long-horizon tasks. This improvement is attributed to the dataset augmentation process, which exposes the SLM to a diverse range of real-world phrasing and thus enhances its robustness to human tone task instructions. 

Compared with the original Llama-3.1-8B, Ro-SLM significantly improves it from being incapable to LLM-like performance in all settings. While Ro-SLM has a relatively lower performance compared with the Corrective GSCE on the Advanced tasks, it is designed to balance its reasoning capability and onboard deployability for resource-constrained environments. The SOTA performance of Corrective GSCE is achieved by its iterative refinement strategy that requires three LLMs. Integrating such a strategy, even with SLMs, will exceed the capability of onboard units of most small robot platforms. 

Overall, these results demonstrate that prompting strategies effective for LLMs cannot transfer to SLMs. By distilling LLM's reasoning and knowledge into SLM, Ro-SLM can substantially improve the reasoning capability and robustness of SLMs and enable SLM to approach the performance of LLM on robotic task planning and code generation.

\begin{table*}[t] 
    \centering
    \begin{tabular}{l | c c | c c}
        \toprule
        \multicolumn{1}{c}{ } & \multicolumn{2}{c}{SR} & \multicolumn{2}{c}{Completeness} \\
         \specialrule{.8pt}{.5pt}{2.5pt} 
       \textbf{Advanced} & GSCE~\cite{GSCE} & Ro-SLM & GSCE~\cite{GSCE} & Ro-SLM \\
        \midrule
        Low & $100\%$ & $66.7\%$ & $100\%$ & $87.9\%$\\
        Medium & $88.9\%$ & $77.8\%$ & $97.8\%$& $86.9\%$\\
        High & $75.0\%$ & $75.0\%$ & $88.9\%$& $75.6\%$\\
        Very High & $25.0\%$ & $50.0\%$ & $73.7\%$& $81.7\%$\\

        \specialrule{.8pt}{.5pt}{2.5pt} 
        \multicolumn{2}{l}{\textbf{Real-world Mapping}} \\
        \midrule
        Low & $100\%$ & $66.7\%$ & $100\%$& $87.9\%$\\
        Medium & $88.9\%$ & $88.9\%$ & $98.9\%$& $95.6\%$\\
        High & $75.0\%$ & $75.0\%$ & $91.7\%$& $72.6\%$\\
        Very High & $25.0\%$ & $50.0\%$ & $84.8\%$& $81.7\%$\\
        \bottomrule
    \end{tabular}
    \caption{Comparison of SR \& Completeness on different complexities: advanced tasks and their real-world mapping.}
    \label{tab:SR complexity}
\end{table*}


\subsubsection{Ablation Study}

An ablation study is performed to analyze the contribution of each component in our Ro-SLM framework. Specifically, we construct variants of Ro-SLM that retain code comments \textit{(Ro-SLM with Comments)}, omit augmentation \textit{(Ro-SLM without Augmentation)}, reduced task diversity \textit{(Ro-SLM Simple Task Only or Complex Task Only)}, and remove GRPO optimization \textit{(Ro-SLM)}. The results are presented in Tables~\ref{tab:overall} and~\ref{tab:overall modified}.

\textbf{Comments} Although the SLM fine-tuned on datasets that retain code comments achieves competitive performance on the Basic tasks, its SR is reduced by $20\%$ on the Advanced tasks. While comments generated by LLMs are useful for human interpretation, they introduce diverse linguistic patterns. For the basic tasks, where the code is short, the comments did not affect. However, Advanced tasks involve longer action sequences, and code comes with more extensive comments. Since SLMs are sensitive to the syntactic patterns of the training data~\cite{SLM-sensitivity}, this variability introduces noise that degrades performance. Accordingly, we remove all code comments in our synthetic dataset to improve the robustness of the fine-tuned SLM.

\textbf{Augmentation} Augmentation improves both SR and Completeness by at least $2\%$ on Basic tasks, and more than $5\%$ on Advanced tasks. These gains indicate that exposing the SLM to more diverse and naturalistic task instructions during training enhances its robustness in both few-stage and long-horizon reasoning. Therefore, we incorporate dataset augmentation to improve the SLM's generalization across varying instruction styles.

\textbf{Task Diversity} As shown in the results, the SLM performs well on Basic tasks but poorly on Advanced tasks when trained exclusively on a simple-task-only dataset, and exhibits the opposite behavior when trained only on complex tasks. This pattern indicates that SLMs adapt to the distribution of the fine-tuning data and tend to overfit to a narrow range of task distribution. In contrast, SLMs trained on a diverse distribution of tasks exhibit more robust performance across varying task types. These results demonstrate the importance of incorporating multiple task-complexity distributions in the dataset synthesis process to ensure the generalization and robustness of fine-tuned SLMs. 

\textbf{Optimization} Removing the optimization step does not affect performance on the Basic tasks, but it reduces the Completeness on the Advanced tasks by 0.6\%. This suggests that GRPO improves step-level reasoning in long-horizon tasks by refining intermediate decision-making, as the reward function encourages step-wise equivalence with the ground-truth code during optimization. Consequently, we incorporate optimization in our framework to enhance SLM's intermediate step reasoning.

\subsection{Analysis over Task Complexity}

We next analyze the performance of the SLM and the LLM on the Advanced tasks across the four complexity levels~\cite{CLGSCE}, as shown in Table~\ref{tab:SR complexity}.

\textbf{SR} As task complexity increases, the SR of the LLM decreases with the increment of complexity, whereas SLM exhibits a different trend. On the original Advanced tasks, the SLM underperforms the LLM at the \textit{Low} level, performs comparably at \textit{Medium} and \textit{High} levels. Notably, the SLM outperforms the LLM under \textit{Very High} level. The improvement is attributed to the availability of the accurate ground-truth code provided through human assistance during dataset construction, which enables the SLM to learn correct complex long-horizon behaviors. These results indicate that the SLM effectively learns mapping tasks to code from the dataset and generalizes to varying task complexities.

\textbf{Completeness} The SLM approaches the performance of LLM across the \textit{Low}, \textit{Medium}, and \textit{High} levels. Although the SLM substantially outperforms the LLM at the \textit{Very High} level in SR, its Completeness remains similar to the LLM. This suggests that while the LLM fails to correctly reason about critical steps in highly complex tasks, the SLM is able to handle these steps more correctly, thereby improving task success.

Overall, although SLMs have inherently more limited reasoning capabilities~\cite{LLM-scaling}, these results demonstrate that Ro-SLM substantially enhances the ability of SLMs to perform complex and long-horizon reasoning robotic tasks at a level comparable to LLM.

\begin{table*}[h] 
    \centering
    \begin{tabular}{l c c}
        \toprule
        & SR & Completeness\\
        \midrule
        Corrective GSCE~\cite{LLM-simulator} & $87.5\%$ & $97.2\%$\\
        Ro-SLM & $75\%$ & $81.1\%$ \\
        \bottomrule
    \end{tabular}
    \caption{Results on ground vehicle.}
    \label{tab:ground vehicle}
\end{table*}

\subsection{Experiment on Ground Vehicle}
To further evaluate the generalization capability of our framework, we extend Ro-SLM to a different robotic platform, a ground vehicle with different APIs and constraints for robot operations. Following the same pipeline, the framework first synthesizes task instructions, generates corresponding ground-truth code, and performs dataset augmentation with respect to the ground vehicle's operations. The trained SLM is then evaluated on eight ground-vehicle operation tasks adopted from~\cite{LLM-simulator}, which control a ROSMASTER X3 to perform operations requiring complex reasoning. As shown in the Table~\ref{tab:ground vehicle}, by leveraging Ro-SLM, the SLM approaches the LLM performance on ground vehicle operations. This suggests that Ro-SLM captures robot task planning rather than overfitting to platform-specific APIs, and demonstrates its potential to generalize across heterogeneous robotic systems.

\section{Conclusion}
In this paper, we present Ro-SLM, a framework that distills LLMs' reasoning and knowledge into SLM to enable reliable task planning and code generation on resource-constrained robot platforms. Our framework first synthesizes a high-quality instruction-code dataset by utilizing LLMs in task instruction generation, corrective code grounding with minimal human assistance, and augmentation. Then the SLM is fine-tuned on the synthetic dataset and further optimized to the best result with LLM-aided optimization. Experiments on UAV and ground vehicle operation tasks show that the SLM in our framework substantially outperforms the original SLM and approaches the LLM-level performance across varying task complexities.

\section*{Limitations}

Despite the promising results demonstrated by our framework, several limitations could be addressed in future research:

\textbf{Generalization and Real-World Deployment} The dataset synthesis is highly task-specific, and transferring our framework across different domains remains challenging. Moreover, although Ro-SLM performs well in a simulated environment, its robustness in real-world robotic systems with unpredictable environmental variables requires further validation.

\textbf{Computational Scalability} Although Ro-SLM substantially improves the performance of Llama-3.1-8B, the model still requires approximately 16GB of memory, which limits deployment on compact edge devices with limited memory capacity. While smaller SLMs (e.g., 3B or 1B models) would fit on a broader range of devices, their performance within the Ro-SLM framework remains to be explored.

\textbf{Human Involvement in Data Generation} While code generation for simple tasks can be automated, minimal human assistance is still needed to verify and correct ground-truth code for high-complexity tasks. This process requires domain expertise and introduces manual overhead, which limits the scalability of dataset construction to new domains and to larger or more complex datasets.

\textbf{Lack of Iterative Correction}. Unlike LLM-based approaches such as Corrective GSCE~\cite{LLM-simulator} that employ multi-round refinement, Ro-SLM relies on a single-pass generation at inference time. While this design improves efficiency and deployability, it limits the model’s ability to correct intermediate reasoning errors, particularly in complex multi-step tasks.

In summary, these limitations suggest that while Ro-SLM provides a strong foundation for enabling onboard language-driven robot control, further work will focus on improving scalability, robustness, and adaptability across diverse real-world robotic platforms.

\section*{Ethical Considerations}
As language model-driven robotic systems become increasingly integrated into real-world deployment, the proposed framework also raises ethical considerations in safety. The generated code controls robotic operation and therefore carries risks of physical harm if improperly deployed. Errors in execution could result in collisions, property damage, or injury to humans. Transferring to real-world robots should include additional safeguards, such as an emergency stop.

\section*{Acknowledgments}
This work was supported by the US National Science Foundation awards 2318710 and 2318711, and the UMass Dartmouth Internal Research Seed Funding Program.


\bibliography{custom}

\appendix

\section{Task Instruction Generation and System Prompts} \label{apdx:instruction sysprompt}
For the training set, we design two distinct system prompts for task-instruction generation with different task complexities. The first prompt induces the LLM to generate 122 tasks containing fewer than five action steps. These tasks are relatively simple and intuitive, and are intended to establish the SLM's basic UAV operation capabilities and short-horizon reasoning skills. The second prompt induces the LLM to generate 42 tasks containing complex, long-horizon UAV operations, such as flying in a series of square patterns. These tasks require multi-stage planning and complex spatial reasoning, and are therefore intended to develop the SLM's long-horizon and logical reasoning capabilities.

For the evaluation set, we modify the number of tasks specified in the system prompts. We generate 28 tasks using the first system prompt and 11 tasks using the second system prompt.

In total, we construct 203 tasks spanning varying lengths, complexity levels, and task types for the fine-tune dataset.

\begin{tcolorbox}[title = System Prompt 1:, colframe=green!70!black]

You are going to generate drone control tasks for me.

\vspace{1em}

Rules for the task generation:

1. The drone tasks involve combinations of world frame movement, rotation, and the drone's body frame movement.

2. The task description must be concise. Do not add additional explanation with "()" in the task.

3. The task description must clearly state the coordinate system (world frame or drone's body frame) for each movement action.

4. The task description must clearly state the rotation direction, and the rotation angle must be divisible by 30.

5. Move and rotate are the only two actions available for the drone.

6. Movement distance should be an integer, the number should be larger than 2 meters and smaller than 10 meters.

7. No more than 5 steps of actions in a task.

8. The task description should be in a human tone.

\vspace{1em}

Here are four example tasks:

1. Fly 3 meters up, then fly 5 meters down in the world frame.

2. Rotate 180 degrees, then fly 5 meters forward in the drone's body frame.

3. Turn to face the local south, then fly 6 meters forward in the drone's body frame.

4. Fly the drone in the top-right direction at an angle of 60 degrees from the horizontal axis, in the YZ plane of the drone's body frame for a distance of 5 meters.

\vspace{1em}

Your output should be tasks only.

\vspace{1em}

Please generate 110 tasks like examples and 12 tasks that fly the drone in XZ or YZ plane like example 4 in the drone's body frame.

\end{tcolorbox}

\begin{tcolorbox}[title = System Prompt 2 Part A:, colframe=green!70!black]

You are going to generate drone control tasks for me.

\vspace{1em}

Rules for the task generation:

1. The drone tasks involve combinations of movement and rotation.

2. The task description must be concise and clear. Do not add additional explanation with "()" in the task.

3. The drone is going to fly a series of square patterns. The patterns involve flying forward, right, backward, and left; flying forward, left, forward, and right; symmetric; reverse; two or more squares; figure of 8.

4. State the purpose, if the task requires a specified facing direction (align, opposite, perpendicular).

5. Move and rotate are the only two actions available for the drone.

6. Movement distance should be an integer, the number should be larger than 2 meters and smaller than 10 meters.

7. The task description should be in a human tone.

\vspace{1em}

Here are four example tasks:

1. Take off and fly up 5 meters. You should fly in a square pattern with 5-meter sides by moving north, east, south, and west in the world axis.
2. Take off and fly up 5 meters. You will examine a square area. You should fly in a square pattern with 5-meter sides by moving forward, left, backward, and right. To examine the square area, the drone should orientate perpendicular to the moving direction on each side of the square. 

3. Take off and fly up 5 meters. You will examine a square area. You should fly in a square pattern with 5-meter sides by moving forward, right, backward, and left. To examine the square area, the drone should orientate perpendicular to the moving direction on each side of the square. Next, ascend another 5 meters and fly the square pattern in reverse order to examine the same area.
\end{tcolorbox}

\begin{tcolorbox}[title = System Prompt 2 Part B:, colframe=green!70!black]
4. Take off and fly up 5 meters. Fly in a square with 5-meter sides. The movement pattern should follow this sequence: forward, right, backward, and left in the world axis. Next, fly a second square that is symmetric with respect to the X-axis in the XY plane. To examine the two square areas, the drone should orientate perpendicular to the moving direction on each side of the squares. 

5. Take off and fly up 5 meters. Fly a figure of 8 on a flat, horizontal plane with each side of 5 meters. The left square is on your left-rear side, and the right square is on your right-front side. You should begin with the left square by flying left. The right square should start from moving north. Additionally, for the left square, the drone is oriented in the flying direction; for the right square, the drone should examine the area while flying the right square pattern. A good strategy for examining is the drone orientate perpendicular to the moving direction on each side of the right square.

\vspace{1em}

Your output should be tasks only.

\vspace{1em}

Generate 42 tasks according to the flight path and pattern from the examples I gave you. You should cover all the patterns in the examples.

\end{tcolorbox}

\newpage
\section{Task Augmentation System Prompt} \label{apdx:aug sysprompt}

\begin{tcolorbox}[title = System Prompt:, colframe=green!70!black]

You are an assistant for thinking up a real-world drone operation task that implements the same drone moving actions as the given task, such that the task description is more realistic for real-world applications. 
\vspace{1em}

Here are some rules for you:

1. Make sure to state the coordinate system (world frame or drone's body frame) for each movement action in your modified task.

2. Your task must perform the same action as the task I gave you. You must not add any additional actions.

3. You must not introduce misleading descriptions that could be interpreted as additional actions.

4. Your output should be human tone-like and should not use uncommon words. 

5. Must output your modified task in one paragraph.
\vspace{1em}

Here is an example:

Query: "Fly 5 meters up, then fly 4 meters down."

Answer: "Perform a vertical clearance check near a storage. In the world frame, ascend 5 meters to inspect the upper vent, then descend 4 meters in the world frame to position the drone near the mid-section for a closer look."

\end{tcolorbox}

\newpage
\section{Reward Function and System Prompt} \label{apdx:reward sysprompt}

Because code generation for the same task may differ syntactically, direct comparison based on textual similarity is impractical. For example, one generation uses ``position'' as a variable name and another generation uses ``base'', although they both refer to the drone's location, their textual representations differ. We therefore adopt the code interpretation strategy from~\cite{LLM-simulator} to assess whether the SLM-generated code and the ground truth code produce the same operational outcome. If the SLM-generated code fully matches the ground truth behavior, the reward is set to 1; otherwise, it is set to 0.

\begin{tcolorbox}[title = System Prompt:, colframe=green!70!black]

You will compare the intentions of the given drone control code with the ground truth code.
\vspace{1em}

You should focus on the code, not the comments. Because code will be the actual actions of the drone.
\vspace{1em}

Here are the available functions for the drone when you interpret the code.

aw.takeoff() - takes off the drone.

aw.land() - lands the drone.

aw.fly\_to([x, y, z]) - flies the drone to the position specified as a list of three arguments corresponding to world XYZ coordinates. The flying speed is 2 meters per second.

aw.get\_yaw() - returns the current yaw of the drone in degrees.

aw.set\_yaw(yaw) - sets the yaw of the drone to the specified value in degrees.

aw.get\_drone\_position() - returns the current position of the drone as a list of 3 floats corresponding to world XYZ coordinates.
\vspace{1em}

Your answer must be 0 or 1, where 0 means the code does not match the ground truth, and 1 means it fully matches.

\end{tcolorbox}

\newpage
\section{Task Dataset Real-World Mapping} \label{apdx:modified dataset}

\begin{table*}
      \begin{tabular}{p{5cm} p{10cm}}
        \toprule
        \multicolumn{1}{c}{\textbf{Task}} & \multicolumn{1}{c}{\textbf{Scenario}} \\
        \midrule
Fly 5 meters up, then fly 4 meters down. & Please perform a vertical inspection of the warehouse wall. In the world frame, ascend 5 meters to check the condition of an overhead ventilation duct, then descend 4 meters in the world frame to inspect a mid-height access panel more closely. \\
\midrule
Fly 5 meters down, then fly 4 meters up. & You are flying under a bridge to inspect structural damage. In the world frame, descend 5 meters to move from the initial hover position down toward the lower girder area for a detailed scan, then ascend 4 meters in the world frame to bring the drone closer to the mid-level of the bridge structure for follow-up imaging. \\
\midrule
Fly 5 meters up, then fly 4 meters up. & You are going to conduct a rooftop antenna inspection at a warehouse. In the world frame, ascend 5 meters to rise above nearby obstacles and reach the general rooftop level, then ascend another 4 meters in the world frame to perform a visual check of the antenna assembly. \\
\midrule
Fly 5 meters down, then fly 4 meters down. & The drone is flying around a tall warehouse. In the world frame, descend 5 meters to move from the overhead inspection position down toward the shelf, then continue descending 4 meters in the world frame to inspect the lower shelves and base of the rack more closely. \\
\midrule
Rotate 180 degrees, then fly 4 meters forward in the drone's body frame. & Perform a corridor inspection. First, in the drone's body frame, rotate 180 degrees to face the opposite direction down the hallway. Then, still in the drone's body frame, fly 4 meters forward to continue the inspection along the newly faced segment of the corridor. \\
\midrule
Rotate 180 degrees, then fly 4 meters backward in the drone's body frame. & The drone is going to conduct a reverse-approach inspection. First, in the drone's body frame, rotate 180 degrees to face the support beams behind the drone, then in the drone's body frame, fly 4 meters backward to carefully reposition closer to the rear side of the structure for detailed imaging. \\
\midrule
Rotate 180 degrees, then fly 4 meters right in the drone's body frame. & The drone is going to fly over the fence of the construction site perimeter. In the drone's body frame, first rotate 180 degrees to turn from facing the inner side to facing the outer fence line, then fly 4 meters to the right in the drone's body frame to laterally reposition along the fence for a side-on visual inspection. \\
\midrule
Rotate 180 degrees, then fly 4 meters left in the drone's body frame. & You are going to inspect the side of the warehouse aisle. First, in the drone's body frame, rotate 180 degrees to face the shelving on the opposite side of the aisle. Then, still in the drone's body frame, fly 4 meters to the left to capture images of the adjacent shelf segment. \\

\midrule
    \end{tabular}
\end{table*}

\begin{table*}
      \begin{tabular}{p{5cm} p{10cm}}
        \toprule

Turn to face the local south, then fly 4 meters forward in the drone's body frame. & The drone is going to perform an inspection of rooftop solar installation along a building's southern edge. First, in the world frame, turn the drone until its front faces local south to align with the panel row, then in the drone's body frame, fly 4 meters forward to approach the center of the southern panel array for closer inspection. \\
\midrule
Turn to face the local south, then fly 4 meters backward in the drone's body frame. & Now, perform an indoor aisle inspection in a warehouse. First, in the world frame, yaw the drone to face local south to align with the aisle direction. Then, in the drone's body frame, fly 4 meters backward to slowly increase the distance from a tall storage rack while maintaining the camera's view of its shelves for documentation. \\
\midrule
Turn to face the local south, then fly 4 meters right in the drone's body frame. & Fly along a rectangular crop field boundary. First, in the world frame, rotate the drone until its front aligns with south to match the survey heading. Then, in the drone's body frame, translate 4 meters to the drone's right side to position it laterally over the field's inner edge for side-view imaging of the boundary. \\
\midrule
Turn to face the local south, then fly 4 meters left in the drone's body frame. & Perform a short side-clearance survey along a building. First, in the world frame, rotate the drone so that its forward-facing camera points toward the local south. Then, move 4 meters to the left in the drone's body frame to inspect the adjacent wall section while keeping the same south-facing view. \\
\midrule
Turn 90 degrees clockwise, then fly 4 meters forward in the drone's body frame. & You are going to inspect the side of a building. First, in the world frame, rotate the drone 90 degrees clockwise to face the adjacent wall. Then, in the drone's body frame, fly 4 meters forward to approach the wall and capture detailed images of the wall. \\
\midrule
Turn 90 degrees clockwise, then fly 4 meters backward in the drone's body frame. & Conduct an indoor aisle inspection in a warehouse. First, in the world frame, rotate the drone 90 degrees clockwise to align it with a side aisle. Then, in the drone's body frame, fly 4 meters backward to safely reverse along that aisle while keeping the camera oriented toward the shelves being inspected. \\
\midrule
Turn 90 degrees clockwise, then fly 4 meters right in the drone's body frame. & The drone is inspecting the side of a house. First, in the world frame, yaw the drone 90 degrees clockwise to align its front camera with the side wall. Then, in the drone's body frame, fly 4 meters to the right to scan along the wall section while maintaining the same distance from the structure. \\
\midrule
Turn 90 degrees clockwise, then fly 4 meters left in the drone's body frame. & The drone is inspecting the side of a wall while maintaining a fixed position relative to the structure. First, in the world frame, rotate the drone 90 degrees clockwise so its camera faces along the wall. Then, in the drone's body frame, translate 4 meters to the left along the wall and capture images of the adjacent wall section. \\
\midrule

  \end{tabular}
\end{table*}

\begin{table*}
      \begin{tabular}{p{5cm} p{10cm}}
        \toprule

Turn to face the local east, then fly 4 meters forward in the drone's body frame. & Conduct a short inspection along a known east-west power line. In the world frame, rotate the drone points toward the local east to align with the line's direction, then in the drone's body frame, fly 4 meters forward along its nose direction to advance to the next inspection point. \\
\midrule
Turn to face the local east, then fly 4 meters backward in the drone's body frame. & Perform an alignment maneuver during a perimeter survey of a construction site. In the world frame, yaw the drone until its nose points toward local east to match the planned heading, then in the drone's body frame, fly 4 meters backward to fly away. \\
\midrule
Turn to face the local east, then fly 4 meters right in the drone's body frame. & Perform a short inspection along a house wall. First, in the world frame, yaw the drone to face local east so its camera aligns with the wall's length, then, maintaining the drone's heading, move 4 meters to the right in the drone's body frame to scan along the wall while keeping its orientation fixed. \\
\midrule
Turn to face the local east, then fly 4 meters left in the drone's body frame. & Inspect the exterior of a wall. First, in the world frame, yaw to align the drone so its nose points toward the local east to face the wall directly. Then, maintaining altitude and orientation, fly 4 meters to the left in the drone's body frame to laterally scan along the wall for structural damage. \\
\midrule
Turn 90 degrees counterclockwise, then fly 4 meters forward in the drone's body frame. & Perform a corridor inspection inside a warehouse. In the world frame, first yaw 90 degrees counterclockwise to align the drone with the inspection aisle. Then, in the drone's body frame, fly 4 meters forward along the aisle to approach the designated inspection point. \\
\midrule
Turn 90 degrees counterclockwise, then fly 4 meters backward in the drone's body frame. & Conduct an indoor inspection for a bookshelf. First, in the world frame, turn the drone 90 degrees counterclockwise to align its camera with a side shelf row. Then, in the drone's body frame, fly 4 meters backward along its longitudinal axis to slowly increase the distance from the shelf while maintaining the same viewing angle for documenting the entire rack. \\
\midrule
Turn 90 degrees counterclockwise, then fly 4 meters right in the drone's body frame. & Perform an indoor aisle inspection in a warehouse. First, in the world frame, rotate the drone 90 degrees counterclockwise to align its forward-facing camera with a perpendicular storage aisle. Then, in the drone's body frame, translate 4 meters to the right to laterally shift along the aisle and inspect the shelf sections on that side. \\
\midrule
Turn 90 degrees counterclockwise, then fly 4 meters left in the drone's body frame. & Perform indoor flying in a shop. First, in the world frame, yaw the drone 90 degrees counterclockwise to align its front with the target shelving row. Then, in the drone's body frame, translate 4 meters to the left to move laterally along the aisle and capture side-view images of the shelf contents. \\
\midrule

\end{tabular}
\end{table*}

\begin{table*}
      \begin{tabular}{p{5cm} p{10cm}}
        \toprule

Turn to face the local west, then fly 4 meters forward in the drone's body frame. & Inspect a solar panel on a rooftop. In the world frame, rotate the drone's nose points toward the local west to align with the panel row, then, in the drone's body frame, fly 4 meters forward along its current heading to move along the row for a closer inspection. \\
\midrule
Turn to face the local west, then fly 4 meters backward in the drone's body frame. & You are flying around a barn. First, in the world frame, yaw the drone to face local west so its camera can align with the target wall. Then, in the drone's body frame, fly 4 meters backward to increase standoff distance from the wall while keeping the same heading for stable visual inspection. \\
\midrule
Turn to face the local west, then fly 4 meters right in the drone's body frame. & Inspect a solar farm row alignment. First, in the world frame, rotate the drone until it faces local west to align with the panel rows. Then, in the drone's body frame, fly 4 meters to the right to shift over to an adjacent lane while maintaining the same heading. \\
\midrule
Turn to face the local west, then fly 4 meters left in the drone's body frame. & Inspect the side of a wall. In the world frame, yaw the drone to face local west to align its camera with the wall, then in the drone's body frame, translate 4 meters to the left to scan along the wall while maintaining the same orientation. \\
\midrule
Fly the drone in the top-right direction at an angle of 30 degrees from the horizontal axis, in the YZ plane of the drone's body frame for a distance of 10 meters. & Inspect the structural integrity of a roof corner joint. In the drone's body frame, in the YZ plane, move the drone 10 meters in the top-right direction at an angle of 30 degrees from the horizontal axis (within that YZ plane) so the onboard camera can capture the upper-right edge of the wall-ceiling junction for detailed imaging. \\
\midrule
Fly the drone in the top-left direction at an angle of 30 degrees from the horizontal axis, in the YZ plane of the drone's body frame for a distance of 10 meters. & Survey the upper-left corner of a warehouse shelf for inventory inspection. In the drone's body frame, move in the YZ plane by flying 10 meters in a direction 30 degrees above the horizontal axis toward the top-left relative to the drone's forward-facing orientation, I would like to see what on the path. \\
\midrule
Fly the drone in the bottom-right direction at an angle of 30 degrees from the horizontal axis, in the YZ plane of the drone's body frame for a distance of 10 meters. & Inspect a vertical cable fastened on a building corner. In the drone's body frame, move 10 meters in the bottom-right direction within its YZ plane at an angle of 30 degrees from the horizontal axis to follow the slanted cable. \\
\midrule
Fly the drone in the bottom-left direction at an angle of 30 degrees from the horizontal axis, in the YZ plane of the drone's body frame for a distance of 10 meters. & Inspect a shelf corner for structural damage: starting from a hover near the aisle, fly the drone in the drone's body frame YZ plane by moving 10 meters in the bottom-left direction relative to its forward-facing orientation, specifically at an angle of 30 degrees downward from the horizontal axis in that YZ plane, so that it can scan along the lower-left edge of the shelving for cracks or deformation. \\
\midrule

  \end{tabular}

\end{table*}

\begin{table*}
      \begin{tabular}{p{5cm} p{10cm}}
        \toprule

Rotate 180 degrees, then fly the drone in the top-right direction at an angle of 30 degrees from the horizontal axis, in the YZ plane of the drone's body frame for a distance of 10 meters. & Inspect the rear side of a cell-tower panel. First, in the drone's body frame, rotate 180 degrees to face the opposite side of the tower. Then, still in the drone's body frame, fly 10 meters in the YZ plane in the top-right direction, maintaining a 30-degree angle from the horizontal axis, to position the drone optimally for capturing detailed images of the rear cabling and connectors. \\
\midrule
Rotate 180 degrees, then fly the drone in the top-left direction at an angle of 30 degrees from the horizontal axis, in the YZ plane of the drone's body frame for a distance of 10 meters. & Conduct an indoor inspection along a warehouse aisle. First, in the drone's body frame, rotate 180 degrees to face the opposite direction down the aisle. Also, in the drone's body frame, fly 10 meters in the YZ plane in a top-left direction, maintaining a 30-degree angle from the horizontal axis to approach and examine elevated shelving and overhead fixtures on the upper-left side of the aisle. \\
\midrule
Rotate 180 degrees, then fly the drone in the bottom-right direction at an angle of 30 degrees from the horizontal axis, in the YZ plane of the drone's body frame for a distance of 10 meters. & Perform an inspection of the rear-lower corner of a building. First, in the drone's body frame, rotate 180 degrees to face the opposite direction along the original heading. Then, still in the drone's body frame within its YZ plane, fly 10 meters in the bottom-right direction at an angle of 30 degrees from the horizontal axis to reach and survey the lower-right rear junction of the wall and support beam. \\
\midrule
Rotate 180 degrees, then fly the drone in the bottom-left direction at an angle of 30 degrees from the horizontal axis, in the YZ plane of the drone's body frame for a distance of 10 meters. & Perform an internal pipe inspection where the drone needs to reorient and then follow a specific diagonal path. First, in the drone's body frame, rotate 180 degrees to face the opposite direction along the pipe. Then, still in the drone's body frame, fly in the bottom-left direction within the YZ plane at an angle of 30 degrees from the horizontal axis for a distance of 10 meters to inspect a lower-side junction of the pipe. \\
\midrule
Turn 60 degrees clockwise, then fly 10 meters forward in the drone's body frame. & Inspect a communication antenna on a building rooftop. First, in the world frame, rotate the drone 60 degrees clockwise to align its front camera with the antenna array, then, in the drone's body frame, fly 10 meters forward to approach the structure for a detailed visual inspection. \\
\midrule
Turn 60 degrees clockwise, then fly 10 meters backward in the drone's body frame. & Inspect a billboard along a roadside. In the world frame, yaw the drone 60 degrees clockwise to align its camera with the billboard surface, then, in the drone's body frame, fly 10 meters backward to increase distance for a wider inspection shot while keeping the same orientation. \\
\midrule
Turn 60 degrees clockwise, then fly 10 meters right in the drone's body frame. & Perform a perimeter scan around a farm corner. In the world frame, yaw the drone 60 degrees clockwise to align its camera with the side fence, then in the drone's body frame fly 10 meters to the right to fly along the fence. \\
\midrule
Turn 60 degrees clockwise, then fly 10 meters left in the drone's body frame. & Conduct an indoor aisle inspection in abandoned storage. First, in the world frame, rotate the drone 60 degrees clockwise to align its forward direction with the target aisle. Then, in the drone's body frame, translate 10 meters to the left to move laterally along the side of a storage rack while keeping its camera oriented down the aisle. \\
\midrule
  \end{tabular}
\end{table*}

\begin{table*}
      \begin{tabular}{p{5cm} p{10cm}}
        \toprule

Turn 60 degrees counterclockwise, then fly 10 meters forward in the drone's body frame. & I need to inspect a building corner for damage. First, in the world frame, rotate the drone 60 degrees counterclockwise to align its camera with the next inspection segment. Then, in the drone's body frame, fly 10 meters forward along its current heading so I can see the corner damage. \\
\midrule
Turn 60 degrees counterclockwise, then fly 10 meters backward in the drone's body frame. & I need to inspect a building for heat leaks. First, in the world frame, rotate the drone 60 degrees counterclockwise to align its thermal camera with the target wall section. Then, in the drone's body frame, fly 10 meters backward to increase the inspection distance while keeping the camera pointed at the same area for a wider thermal overview. \\
\midrule
Turn 60 degrees counterclockwise, then fly 10 meters right in the drone's body frame. & You are a drone flying inside a warehouse. In the world frame, first yaw the drone 60 degrees counterclockwise to align its forward-facing camera toward the new shelf row, then in the drone's body frame, translate 10 meters to the right to move laterally along the shelf for a continuous side-view inspection of stored items. \\
\midrule
Turn 60 degrees counterclockwise, then fly 10 meters left in the drone's body frame. & I need to see the status of my roof panel. First, in the world frame, yaw the drone 60 degrees counterclockwise to face along the panel row, then in the drone's body frame, fly 10 meters to the left to position the drone over the adjacent panel for next panel imaging. \\
\bottomrule

          \end{tabular}
          \caption{Application Scenarios in Basic Task Dataset}
    \label{tab:basic task mapping}
\end{table*}

\newpage

\begin{table*}
  
      \begin{tabular}{p{5cm} p{10cm}}
        \toprule
        \multicolumn{1}{c}{\textbf{Task}} & \multicolumn{1}{c}{\textbf{Scenario}} \\
        \midrule
Take off and fly up 5 meters. You should fly in a square pattern with 5-meter sides by moving forward, right, backward, and left along the world axis. &  The drone is going to carry out an inspection flight above an empty parking lot. First, take off and ascend 5 meters in the world frame to reach a safe inspection height. Then, in the world frame, fly in a square pattern with 5-meter sides: move 5 meters forward to scan the area ahead, 5 meters to the right to check the neighboring section, 5 meters backward to pass over the starting area again, and finally 5 meters to the left to return to the original position at the same altitude. \\
\midrule
Take off and fly up 5 meters. You should fly in a square pattern with 5-meter sides by moving forward in the drone's body frame and turning right at each corner. &  Control the drone to conduct an automated perimeter check around a small farm. First, take off and climb vertically 5 meters in the world frame to reach the inspection altitude. Then, in the drone's body frame, fly forward 5 meters to begin the pass along one side, make a right turn to face the next side, fly forward another 5 meters, and repeat this pattern of turning right and flying forward 5 meters at each corner until completing a square path with 5-meter sides.\\
\midrule
Take off and fly up 5 meters. You should fly in a square pattern with 5-meter sides by moving forward, right, backward, and left along the world axis. Make sure the drone is oriented in the flying direction. & Carry out a simple perimeter check around a small equipment zone. First, take off and climb vertically 5 meters in the world frame. Then, keeping the drone at this height and always yawed so its front faces the direction of travel, fly a square path with 5-meter sides in the world frame by moving 5 meters forward, then 5 meters to the right, then 5 meters backward, and finally 5 meters to the left to return to the starting point. \\
\midrule
Take off and fly up 5 meters. You should fly in a square pattern with 5-meter sides by moving forward, right, backward, and left along the world axis. Make sure the drone is oriented opposite to the flying direction. & I need the drone to fly over a rooftop area. First, take off and ascend 5 meters in the world frame to reach a safe inspection height. Then, fly a square survey pattern with 5-meter sides in the world frame: move forward 5 meters, then right 5 meters, then backward 5 meters, and then left 5 meters to return to the starting horizontal position. Throughout the entire square flight, keep the drone's heading opposite to its current direction of travel so that its camera is always facing the area already inspected. \\
\midrule
Take off and fly up 5 meters, then turn to face west. Next, turn to face east. From your current position, fly in a square with 5-meter sides while continuously facing east. The movement pattern should follow this sequence: north, east, south, and west along the world axis. & Please conduct a short positioning and orientation test near a building. First, take off and ascend 5 meters in the world frame, then yaw to face west to verify the heading, followed by a yaw to face east to set the final inspection direction. From this fixed altitude and always keeping the drone's nose facing east, fly a 5-meter square path in the world frame, moving north 5 meters, then east 5 meters, then south 5 meters, and finally west 5 meters to return to the starting horizontal position. \\
\midrule

      \end{tabular}

\end{table*}

\begin{table*}
      \begin{tabular}{p{5cm} p{10cm}}
        \toprule
Take off and fly up 5 meters. You will examine a square area. You should fly in a square pattern with 5-meter sides by moving forward, right, backward, and left. To examine the square area, the drone should orientate perpendicular to the moving direction on each side of the square. & Perform a low-altitude inspection of a construction site. First, take off and ascend vertically 5 meters in the world frame to reach the inspection height. Then, survey a square 5-meter-by-5-meter area by flying in a square path: move 5 meters forward, then 5 meters right, then 5 meters backward, and finally 5 meters left, all in the same world frame. While flying along each side of the square, keep the drone's body orientation perpendicular to its current direction of motion so its camera faces sideways relative to the flight path during the entire inspection. \\
\midrule
Take off and fly up 5 meters. You will examine a square area. You should fly in a square pattern with 5-meter sides by moving forward, right, backward, and left. To examine the square area, the drone should orientate perpendicular to the moving direction on each side of the square. Next, ascend another 5 meters and fly the square pattern in reverse order to examine the same area. & The drone carries out a two-level aerial inspection of warehouse shelves. First, take off and climb 5 meters in the world frame. At this height, survey a 5-by-5-meter square by flying in a square pattern: move 5 meters forward, then 5 meters right, then 5 meters backward, then 5 meters left, keeping the drone's heading perpendicular to its direction of motion for each leg of the square. After completing this loop, ascend another 5 meters in the world frame and repeat the same square pattern in the reverse order to inspect the same area from a higher altitude. \\
\midrule
Take off and fly up 5 meters. Fly in a square with 5-meter sides. The movement pattern should follow this sequence: forward, right, backward, and left along the world axis. Next, fly a second square that is symmetric with respect to the X-axis in the XY plane. To examine the two square areas, the drone should orientate perpendicular to the moving direction on each side of the squares. & Conduct an aerial calibration around a building rooftop. First, take off and ascend 5 meters in the world frame. At that altitude, survey a square inspection area with 5-meter sides by flying straight forward 5 meters, then right 5 meters, then backward 5 meters, and then left 5 meters, all defined in the world frame. Next, inspect a second square area that is the mirror image of the first one across the world X-axis in the XY plane, following the same square path. Throughout both square inspections, keep the drone's heading perpendicular to its current flight direction along each side, so the camera consistently views sideways relative to the motion. \\
\midrule
Take off and fly up 5 meters. Fly a figure of 8 on a flat, horizontal plane with each side of 5 meters. The left square is on your left-rear side, and the right square is on your right-front side. You should begin with the left square by flying left. The right square should start from moving forward. Make sure the drone is oriented in the flying direction for all squares. & The drone is a trainer; it is going to fly a training flight for inspecting two adjacent rooftop sections. First, take off and climb 5 meters in the world frame to reach a safe inspection height. Then, at this constant altitude, trace a flat “figure-8” path made of two 5-meter squares on the horizontal plane: the first square is located to your left rear side and is flown by moving left first, and the second square is located to your right front side and is flown by moving forward first. Throughout the maneuver, keep the drone's nose pointed in the direction of motion along each segment of both squares. \\
\midrule

      \end{tabular}

\end{table*}

\begin{table*}
      \begin{tabular}{p{5cm} p{10cm}}
        \toprule
Take off and fly up 5 meters. Fly a figure of 8 on a flat, horizontal plane with each side of 5 meters. The left square is on your left-rear side, and the right square is on your right-front side. You should begin with the left square by flying left. The right square should start from moving north. Additionally, for the left square, the drone is oriented in the flying direction; for the right square, the drone should examine the area while flying the right square pattern. A good strategy for examining is the drone orientate perpendicular to the moving direction on each side of the right square. & Conduct an aerial inspection around two adjacent houses. First, take off and ascend vertically 5 meters in the world frame to reach the inspection height. Then, at this constant altitude, fly a flat figure-of-eight pattern made of two adjacent 5-meter-by-5-meter squares in the horizontal plane. The first (left) square is positioned to your left-rear side, and you start this square by moving left while keeping the drone's nose aligned with the direction of motion along each side. After completing the left square, continue directly into the second (right) square, which is positioned to your right-front side and starts with motion toward the north. While flying the right square, maintain the same 5-meter side lengths and altitude, but use the drone to examine the surrounding area by keeping its head oriented perpendicular to the direction of travel on each side, as if scanning the environment sideways while following the square path. \\
\midrule
Take off and climb up 5 m. Fly a 5 m square by moving forward, left, backward, and right. Then fly a 5 m square by moving forward, right, backward, and left. & Perform a basic inspection pattern around a small equipment zone. First, take off and climb straight up 5 meters in the world frame to reach a safe inspection height. Then fly a 5-meter square path in the world frame by moving 5 meters forward, 5 meters left, 5 meters backward, and 5 meters right, returning to the starting point above the equipment. After that, repeat a 5-meter square scan in the world frame but this time move 5 meters forward, 5 meters right, 5 meters backward, and 5 meters left, again returning to the same hovering point. \\
\midrule
Take off and climb up 5 m. Fly a 5 m square (forward/right/back/left). Then shift 5 m south, and fly the same square pattern again. & Inspect a rooftop section of a warehouse. In the world frame, take off and climb up 5 meters to reach inspection altitude, then fly a 5-meter square path by moving 5 meters forward, 5 meters right, 5 meters backward, and 5 meters left to scan the first area. After finishing this square, shift 5 meters south in the world frame to align over the next section of the roof, and repeat the same 5-meter square path with the same forward, right, back, and left movements to complete the second area scan. \\
\midrule
Take off and climb up 5 m. Fly a 5 m square (forward/right/back/left). And fly the same square pattern again. Make sure the drone is oriented to the flying direction in each square. & The drone is going to perform a simple flight check in an open test field. First, take off and climb up 5 meters in the world frame to reach a safe inspection height. Then have the drone fly a square path with 5-meter sides in the world frame by moving forward, then right, then back, then left, keeping its heading aligned with the current flying direction along each side of the square. After completing this square once, repeat the same 5-meter square pattern in the same way, again making sure the drone's orientation always matches the direction of travel on each leg. \\
\midrule

      \end{tabular}

\end{table*}

\begin{table*}
      \begin{tabular}{p{5cm} p{10cm}}
        \toprule
Take off and fly up 5 meters. You should fly in a square pattern with 5-meter sides by moving forward, right, backward, and left along the world axis. Then fly the same square pattern again. Make sure the drone is oriented opposite to the flying direction. & Inspect a rooftop ventilation area with repeated coverage at a fixed height. In the world frame, take off and ascend 5 meters to reach the inspection altitude. At this height, fly a square patrol route with 5-meter sides by moving forward, then right, then backward, then left in the world frame, keeping the drone's heading always opposite to its current direction of motion during this square. After completing the first square, repeat the same square pattern a second time at the same altitude, again ensuring the drone stays oriented opposite to its flying direction for each leg of the path. \\
\midrule
Take off and fly up 5 meters, then turn to face south. Next, turn to face east. From your current position, fly in a square with 5-meter sides while continuously facing east. The movement pattern should follow this sequence: north, east, south, and west in the world axis. & We just repaired a drone, and now conduct an outdoor positioning and heading check for a drone. First, take off and climb 5 meters in the world frame to reach a safe testing height, then rotate in place to face south, and then rotate again in place to face east to confirm heading control. From this fixed starting point, fly a 5-meter-side square path in the world frame while always keeping the drone's nose pointing east, following straight segments in this order: move north 5 meters, then east 5 meters, then south 5 meters, and finally west 5 meters to return to the starting position. \\
\midrule
Take off and climb up 5 m. Fly a 5 m square (forward/right/back/left). Then shift 5 m north, and fly the same square pattern again. The drone will examine the second square area, where the drone should orientate perpendicular to the moving direction on each side of the second square. & You are going to survey in two neighboring grid cells. First, take off and climb 5 meters in the world frame. From there, fly a 5-meter square path by moving 5 meters forward, then 5 meters right, 5 meters back, and 5 meters left in the world frame to complete the first area scan. Next, shift 5 meters north in the world frame to reach the second grid cell, and repeat the same 5-meter square pattern (forward, right, back, left). During this second square, keep the drone's heading perpendicular to its motion along each side so that it is rotated 90 degrees relative to the direction of travel while it examines this second area. \\
\midrule
Take off and fly up 5 meters. You will examine a square area. You should fly in a square pattern with 5-meter sides by moving forward, left, backward, and right. To examine the square area, the drone should orientate perpendicular to the moving direction on each side of the square. & Carry out a short aerial inspection over a small field. First, take off and climb vertically 5 meters in the world frame to reach a safe inspection height. Then, at this fixed altitude, survey a square area with 5-meter sides by flying in a square path: move 5 meters forward, then 5 meters left, then 5 meters backward, and finally 5 meters right in the world frame to return to the starting point. While flying along each side, keep the drone's heading perpendicular to the current flight direction so the camera looks sideways relative to the movement for better coverage of the area. \\
\midrule

      \end{tabular}
\end{table*}

\begin{table*}
      \begin{tabular}{p{5cm} p{10cm}}
        \toprule
Take off and fly up 5 meters. Fly in a square with 5-meter sides. The movement pattern should follow this sequence: forward, right, backward, and left along the world axis. Then descend 5 meters. Next, fly a second square that is symmetric with respect to the north-south(X) axis. You will examine the two square areas, and the drone should orientate perpendicular to the moving direction on each side of the squares. & Conduct an automated inspection of two adjacent rooftop zones. First, in the world frame, take off and ascend 5 meters to the inspection height. Then survey the first zone by flying a square path with 5-meter sides, moving in sequence: forward, right, backward, and left, all defined in world axes. After completing this loop, descend 5 meters in the world frame to return to the starting altitude. Next, perform a second square survey at the same scale, with the square path mirrored about the north-south (X) axis in the world frame, so it is symmetric to the first one. Throughout both square paths, keep the drone's heading oriented perpendicular to its direction of motion along each side to simulate a side-looking inspection payload. \\
\midrule
Take off and fly up 5 meters. Fly a figure of 8 on a flat, horizontal plane with each side of 5 meters. The first square of 8 is on your left-front side, and the right square is on your right-rear side. You should begin with the first square by flying left. The right square should start from moving south. Make sure the drone is oriented in the flying direction for all squares. & You are responsible for conducting an inspection flight around a small neighborhood. First, take off and climb vertically 5 meters in the world frame to reach a safe inspection height. Then, at that constant altitude, trace a horizontal figure 8 pattern with two adjoining 5-meter-wide squares in the world frame: fly the first 5-meter square to your left-front side, starting by moving left, keeping the drone's nose always pointed along the direction of motion; then continue directly into the second 5-meter square on your right-rear side, starting that square by moving south in the world frame, again keeping the drone oriented along the direction of travel for every segment of both squares. \\
\midrule
Take off and fly up 5 meters. Fly a figure of 8 on a flat, horizontal plane with each side of 5 meters. The first square of 8 is on your left-front side, and the right square is on your right-rear side. You should begin with the first square by flying left. The right square should start from moving south. Additionally, for the left square, the drone is oriented in the flying direction; for the right square, the drone should examine the area while flying the right square pattern. A good strategy for examining the inward square area is to let the drone orientate perpendicular to the moving direction on each side of the right square. &  You should control the drone to do a low-altitude inspection of a small construction site. First, take off and ascend vertically 5 meters in the world frame. Then, at that height, survey two adjacent 5 m*5 m square zones on a flat horizontal plane in a figure 8 path: the first square lies to your left front and the second to your right rear. Start by flying the left-front square, beginning with a 5-meter segment to the left and completing the full square while keeping the drone's nose pointed along the direction of travel on each side. After finishing that square, continue into the right-rear square, starting its path with a segment heading south and flying the same 5-meter-per-side square. During this right-rear square, inspect the interior of the square by keeping the drone's camera pointed toward the inside area: on each side, orient the drone so its body is perpendicular to the direction of motion while it follows the right-rear square pattern. \\

\bottomrule

      \end{tabular}
      \caption{Application Scenarios in Advanced Task Dataset}
    \label{tab:advaced task mapping}
\end{table*}



\end{document}